# A Profile-Based Binary Feature Extraction Method Using Frequent Itemsets for Improving Coronary Artery Disease Diagnosis

Ali Yavari, Amir Rajabzadeh*, Fardin Abdali-Mohammadi

*Department of Electrical and Computer Engineering, Razi University, Kermanshah, Iran*

ARTICLE INFO

ABSTRACT

Recent years have seen growing interest in the diagnosis of Coronary Artery Disease (CAD) with machine learning methods to reduce the cost and health implications of conventional diagnosis. This paper introduces a CAD diagnosis method with a novel feature extraction technique called the Profile-Based Binary Feature Extraction (PBBFE). In this method, after partitioning numerical features, frequent itemsets are extracted by the Apriori algorithm and then used as features to increase the CAD diagnosis accuracy. The proposed method consists of two main phases. In the first phase, each patient is assigned a profile based on age, gender, and medical condition, and then all numerical features are discretized based on assigned profiles. All features then undergo a binarization process to become ready for feature extraction by Apriori. In the last step of this phase, frequent itemsets are extracted from the dataset by Apriori and used to build a new dataset. In the second phase, the Genetic Algorithm and the Support Vector Machine are used to identify the best subset of extracted features for classification. The proposed method was tested on the Z-Alizadeh Sani dataset, which is one the richest databases in the field of CAD. Performance comparisons conducted on this dataset showed that the proposed method outperforms all major alternative methods with 98.35% accuracy, 100% sensitivity, and 94.25% specificity. The proposed method also achieved the highest accuracy on several other datasets.

## 1. Introduction

Coronary Artery Disease (CAD) is the most common type of disorder of the heart and blood vessels also referred to as cardiovascular disease (CVD). The World Health Organization (WHO) estimates that by 2030, 23.6 million people worldwide will die from CVD [1]. Given the notable costs and health risks of some CAD diagnosis methods (e.g. angiography, which is the main method to diagnose CAD), physicians are turning to machine learning methods to detect this disease. The machine learning methods most commonly used for this purpose are supervised learning or classification algorithms.

One of the most important phases of machine learning before classification is the feature extraction phase. In general, this phase involves deriving a set of features from a set of raw data or existing features usually through a transform. This phase can have a profound impact on the outcome of the classification process by giving it rich discriminative features as input [2]. Many researchers have tried to use deep neural networks (e.g. Convolutional Neural Networks or Recurrent Neural Networks) to automate feature extraction [3-8]. However, a great challenge in this area is that end-to-end deep learning requires having enough data to properly train the deep network [9]. But since in many fields including healthcare, rich datasets with large enough samples are not publicly accessible, it is still common to use manual feature extraction methods or feature generation methods such as Genetic Programming for feature extraction in these fields [10], as they can compete with deep neural networks in terms of classification accuracy.

This paper presents a CAD diagnosis method with a new feature extraction technique called the Profile-Based Binary Feature Extraction (PBBFE). This method involves extracting frequent itemsets with Apriori, which is a well-known algorithm for Association Rule Mining (ARM) and then adding them to the original dataset to be used as features. The challenge with this method is that Apriori only accepts binary features as input, which means all numerical and categorical features must be converted into their binary equivalents, but this is problematic in health databases, where there is no fixed reference value for some features. For example, the normal range of hemoglobin level is 10.0-18.0 g/dl for 15-30 day old newborns, but it is 9-14 g/dl for 30-90 day old babies [11]. To resolve this problem, we use a method called profile-based partitioning, which has been introduced in [12]. In this method, each patient is assigned a profile based on age, gender, and medical conditions (e.g. pregnancy), which is then used to determine the normal range of numerical features for that patient. Then, numerical features are partitioned based on these normal ranges. The use of this profile remarkably improves the feature partitioning accuracy, which helps Apriori find frequent itemsets with higher accuracy. Also, according to [2, 13], using frequent itemsets as features helps increase the classification accuracy, as a feature may have higher discriminative power when combined with others.

\* *Corresponding author.* Tel.: +98-83-34343410
E-mail addresses: ali.yavari@razi.ac.ir (A. Yavari), rajabzadeh@razi.ac.ir (A. Rajabzadeh), fardin.abdali@razi.ac.ir (F. Abdali-Mohammadi)

The proposed method consists of two main phases. The first phase involves assigning profiles, discretizing numerical features, binarizing features, and finally extracting frequent itemset and adding them to the original database. Since this process may result in having too many features some of which may not be useful, in the second phase, the Genetic Algorithm (GA) is used to filter the features. This phase involves building different subsets of features and evaluating them using the Support Vector Machine (SVM) to ultimately find the optimal subset, i.e. the subset that results in having the lowest classification error. The main contributions of this article to the literature are as follows:

1) Using the concept of profile for more accurate extraction of frequent itemsets;
2) Extracting highly discriminative and interpretable binary features with the help of frequent itemsets;
3) Using PBBFE to achieve higher accuracy in CAD diagnosis compared to previous works (In this regard, it should be noted that even 1% higher accuracy in the diagnosis of CAD (and other diseases) may save many lives).

The remainder of this article is organized as follows. A review of previous works in the field of classification-based CAD diagnosis is provided in Section 2. The components of the proposed method and PBBFE are described in Section 3. The results from testing the proposed method on the Z-Alizadeh Sani dataset are reported in Section 4. The performance of the method on other datasets is discussed in Section 5. In the end, a summary of results is given in Section 6.

## 2. Previous work in the field of CAD diagnosis with machine learning techniques

As stated in the first section, many researchers have worked on the diagnosis of diseases using supervised learning algorithms. The most notable examples of research in the field of CVD diagnosis with classification algorithms are Ref. [14-35].

In [19-21, 28, 30, 35-38], researchers developed a number of classification algorithms for CAD diagnosis and tested them on the Z-Alizadeh Sani dataset. In a study by Alizadehsani et al. [19], where they used SMO, Naive Bayes, KNN, and SVM classification algorithms for CAD diagnosis, the best result was achieved with the SMO algorithm, which had an accuracy of 92.09%. In [21], CAD diagnosis was performed with Bagging SMO, Naive Bayes, SMO, and ANN algorithms. This study reported the best algorithms to be SMO and Bagging SMO with an accuracy of 94.08% and 93.40%, respectively. In [20], two algorithms called C4.5 and Bagging were used for CAD diagnosis. In this study, the model built with the Bagging algorithm provided the best results in terms of detecting heart disease. In [30], Arabasadi et al. proposed a hybrid method consisting of ANN and GA for CAD diagnosis. The core idea of this study was to use GA to determine the optimal weights of neurons in the neural network. This method managed to achieve 9.53% higher accuracy than conventional ANN and was reported to have 97% sensitivity and 92% specificity in CAD diagnosis. In [35], three classifiers were used to detect stenosis in the three coronary arteries Left Circumflex (LCX), Right Coronary Artery (RCA), and Left Anterior Descending (LAD) in order to increase the accuracy of CAD diagnosis. The results of this study showed that the method produces excellent outputs with an accuracy of 96.40%, sensitivity of 100%, and specificity of 88.1% when implemented with SVM. Also in [38], a method called C-CADZ was able to produce an accuracy, sensitivity and specificity equal to 97.37%, 98.15% and 95.45%, which is the best method proposed so far.

In [23], Nahar et al. used six algorithms, namely Part, J48, Naïve Bayes, SMO, AdaBoost m1, and IBK to detect heart diseases. In addition to artificial intelligence-based feature selection methods, they used a method called the Medical knowledge motivated Feature Selection (MFS) to identify useful features based on medical knowledge. This study reported that feature selection with MFS improved the classification results. In [27], an ANN with 13 neurons in the input layer, 6 neurons in the hidden layer, and 1 neuron in the output layer was developed for CVD diagnosis. The accuracy of this method was reported to be 85%.

In [17], researchers proposed a model for CAD diagnosis using SVM, which uses the Principle Component Analysis (PCA) method to reduce the dimensionality of features. The results of this study showed that the model built with 18 out of 23 possible features offered the best accuracy, which was 79.17%. In [15], Das et al. used the ensemble approach to develop a model for detecting heart problems. This method involves running three independent multi-layer feed-forward neural networks and then integrating their results using the ensemble approach to reach the final output. The accuracy of this model was reported to be 89.01%.

## 3. Proposed method: CAD diagnosis with PBBFE

This section describes the components of the proposed method. As shown in Figure 1, the proposed method consists of two main phases. The first step of the first phase (profile-based binary feature extraction) is to assign each patient a profile based on their age, gender, and medical condition. In the second step, all numerical features are discretized based on the created profile. The third step is to turn all features into their binary equivalents. In the last step of this phase, all frequent itemsets are extracted by the Apriori algorithm and added to the original dataset. In the second phase (classification with feature selection), GA is used to identify the most useful and discriminative features among the set of all features in the data. This process is done simultaneously with classification (with the SVM algorithm). A more detailed description of these two phases is given in the following subsections.

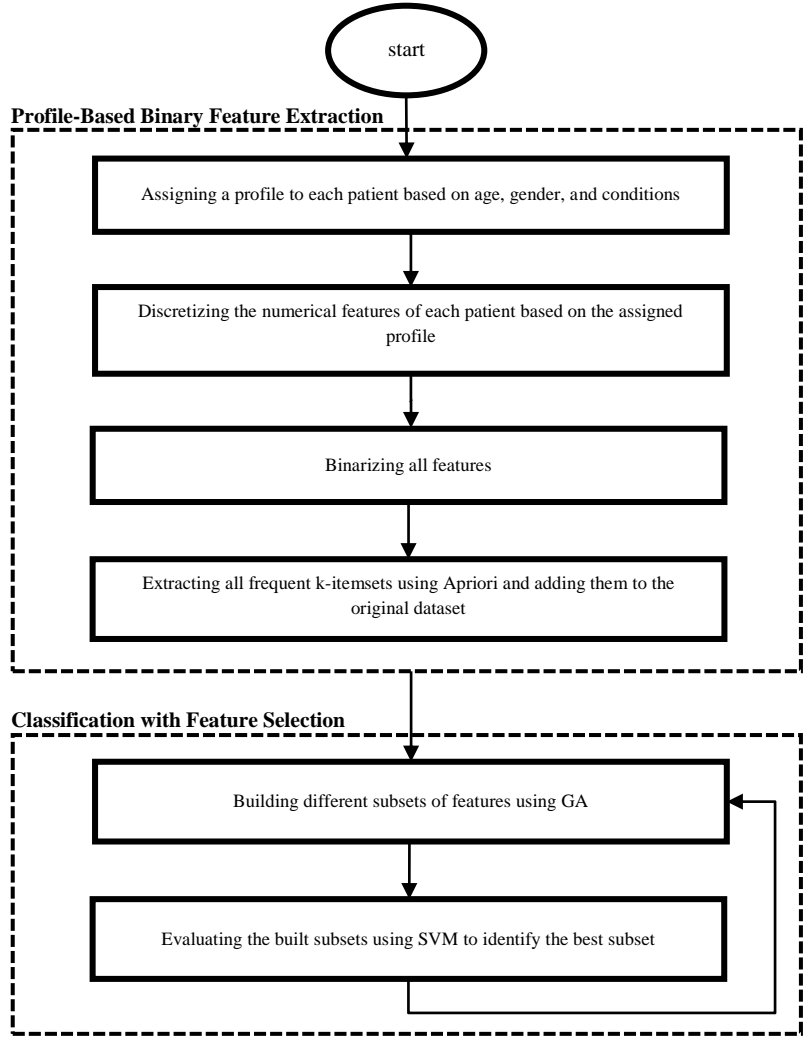

**Fig. 1.** Components of the proposed method; CAD diagnosis with profile-based binary feature extraction.

*3.1. Phase I: Profile-based binary feature extraction*

This section explains the four steps of the first phase of the method, that is, how profiles are created, how numerical features are discretized based on profiles, how features are binarized, and finally how feature extraction is done with Apriori.

*3.1.1. Step 1: Profile creation*

This step involves giving each patient (data record) a profile based on age, gender, and medical condition with the ultimate purpose of improving partitioning accuracy. As explained in [12], assuming that the number of age ranges, genders, and medical conditions by which the reference value of features vary is respectively $\alpha_{age}$, $\alpha_{gender}$, and $\alpha_{condition}$, the total number of profiles that can be created for all patients (num.of.profiles) will be the product of these three numbers, i.e. $\alpha_{age} \times \alpha_{gender} \times \alpha_{condition}$.

*3.1.2. Step 2: Profile-based discretization of numerical features*

After assigning a profile to each patient, all numerical features should be discretized based on the normal ranges given in a credible reference. For this purpose, first, the crisp normal range of each numerical feature for each profile (for example, the normal range of hemoglobin level for pregnant women in the age range of 20-30) must be extracted from a credible reference and then the feature must be discretized based on these normal ranges. Please note that a feature may have two different normal ranges for two different profiles. For example, the normal range of hemoglobin level for a normal middle-aged man (profile $p_i$) will be different from that for a normal middle-aged woman (profile $p_j$) [11, 12].

*3.1.3. Step 3: Binarization of features*

After discretizing features, they must undergo a binarization process. For categorical features, this binarization is done in two ways:

1) For binominal features, the value associated with the presence/occurrence of a disease/condition is labeled True and the alternative value is labeled False. For example, assuming that the feature DM representing diabetes in patients has two values Yes and No, the Yes value will be labeled True and the No value will be labeled False.

2) For polynomial features, the value associated with the presence/occurrence of a disease/condition is labeled True and the alternative values are labeled False. For example, if the feature Cr representing creatinine level in patients has three values Normal, Low, and High, the Low and High values will be labeled True and the Normal value will be labeled False.

*3.1.4. Step 4: Feature extraction with Apriori*

Apriori is one of the most well-known ARM algorithms [39]. The inputs of this algorithm are transactions containing a number of items (each itemset consists of a number of items, and an itemset with k items is called a k-itemset). This algorithm consists of two steps: extracting frequent k-itemsets and extracting association rules from these itemsets. A k-itemset is said to be frequent if it is repeated across the dataset more than (or equal to) a certain number of times, which is specified by the user. This user-specified threshold is called minimum support or Min_sup ($0<$Min_sup$<1$). To use this algorithm in the field of health, it is necessary to consider each patient as a transaction and each feature as an item of this transaction. But to be usable in the algorithm, these items must be binary, which is why binarization was performed in the previous step.

The proposed method only uses the first step of this algorithm for feature extraction. More specifically, the algorithm is used to extract all frequent k-itemsets, which are then injected into the original dataset as features. The Min_sup value has a great impact on the diagnosis accuracy of the proposed method. While using a high Min_sup value will lead to the elimination of potentially useful features, using a low Min_sup value tends to increase the number of useless features and complexity of the classification process. The final output of this phase is a dataset made up of original features and those extracted in this step.

*3.2. Phase II: classification with feature selection (concurrent feature selection-classification)*

The parameter Min_sup can be used to control the number of frequent itemsets generated by Apriori [39]. However, if a dataset has a large number of features, then the number of extracted features will also be large. Thus, it is necessary to embed a feature selection process in the proposed method to filter the features (this applies to all machine learning problems that have to work with a great number of features, as it is crucial for increasing or at least maintaining the classification accuracy). This section first provides a brief description of GA and SVM and then describes the criteria used in the performance evaluation of classification algorithms.

*3.2.1. Genetic Algorithm*

Genetic Algorithm (GA) is one of the most popular and powerful population-based optimization algorithms and has shown great capability in solving both discrete and continuous optimization problems [40]. Thanks to its two powerful operators called Crossover and Mutation, this algorithm has found extensive use in the field of feature selection [41]. In the proposed method, a combination of GA with SVM is used for feature selection. In each iteration of this combined method, Crossover and Mutation operators are used to build different subsets of features and then SVM is used to determine their fitness (classification accuracy). This continues until the stop condition is met (reaching a certain number of iterations, a certain level of classification accuracy, etc.).

*3.2.2. Support Vector Machine (SVM)*

Support Vector Machine (SVM) is a powerful and widely used supervised learning algorithm capable of classifying both linear and nonlinear data [42]. To solve classification problems where the data is not linearly separable, this algorithm uses a nonlinear mapping function called a kernel to take the data to a higher dimension, where it uses the concepts of support vectors and margins to find a hyperplane whereby one class can be linearly separated from another [2]. The kernels most widely used in SVM are polynomial, Radial Basis Function (RBF), and sigmoid.

**Table 1**: Confusion matrix

| Actual\predict | Predicted: Yes | Predicted: No |
|---|---|---|
| **Actual: Yes** | $f_{11}$ (TP) | $f_{10}$ (FN) |
| **Actual: No** | $f_{01}$ (FP) | $f_{00}$ (TN) |

*3.2.3. Performance evaluation criteria for classification models*

One way to visualize the performance of a classification model is to create its confusion matrix, which also helps determine several other evaluation criteria. The structure of this matrix for a binary classification problem is shown in Table 1. Here, $f_{11}$, $f_{01}$, $f_{10}$, and $f_{00}$ denote true positive (TP), false positive (FP), false negative (FN), and true negative (TN), respectively. This matrix can be used to compute several important evaluation criteria, namely accuracy, sensitivity, and specificity, which are given by Equations 1, 2, and 3. To evaluate the performance of a classifier in a binary classification problem, it is common to use a graphical plot known as the Receiver Operating Characteristic (ROC). Typically, the horizontal axis of this plot is the False Positive Rate (FPR) and its vertical axis is the True Positive Rate (TPR), which are given by Equations 4 and 5. The greater the area under the ROC curve (AUC) is for a model, the better is the model.

$$Accuracy = \frac{f_{00} + f_{11}}{f_{11} + f_{00} + f_{01} + f_{10}} \tag{1}$$

$$Sensitiviy = \frac{f_{11}}{f_{11} + f_{10}} \tag{2}$$

$$Specificity = \frac{f_{00}}{f_{00} + f_{01}} \tag{3}$$

$$FPR = \frac{f_{01}}{f_{00} + f_{01}} \tag{4}$$

$$TPR = \frac{f_{11}}{f_{11} + f_{10}} \tag{5}$$

## 4. Test results

This section first provides a description of the main dataset used to test the method (the Z-Alizadeh Sani dataset) and then reports the results of the method on this database.

*4.1. Z-Alizadeh Sani dataset*

The Z-Alizadeh Sani dataset is one of the most recent and up-to-date datasets in the field of CAD. This dataset consists of features collected from 303 patients [21, 28, 30]. Out of all the features contained in this database (54 features in the original version), we used 48 features as input and 1 feature (the patient having CAD) as the target. Out of these features, 18 were numerical and 30 were categorical. A list of these features and the range of their values in the dataset are given in Table 2.

In the first phase of the proposed method, each patient had to be given a profile based on age, gender, and medical condition as explained in Section 3.1.1. As in [12], four profiles of Table 3 were defined for the patients in the dataset. For example, $p_2$ is the profile of normal middle-aged women (given the ranges defined in Table 4, the parameter Age is labeled Normal and High).

In the second step, as explained in Section 3.1.2, the crisp normal range of each feature for each profile was extracted from a credible source. This source was primarily Ref. [11], but Ref. [21] was also used if the normal range of a feature was not available in [11]. Next, numerical features were discretized according to these normal ranges. Table 4 shows the normal range of each numerical feature for different profiles (in this table, discretized features are marked with the subscript "2"). For example, the feature HB2 (discretized hemoglobin) has a normal range of 13.5-17.5 g/dl for the profiles $p_1$ and $p_3$ and a normal range of 12-16 g/dl for the profiles $p_2$ and $p_4$.

Next, all features were binarized as instructed in Section 3-1-3. It should be noted that the feature gender was converted to two binary features: male (True and False) and female (True and False). The final output of this step was a dataset with 49 features and 1 target, all of which were binary.

Table 2: List of numerical and categorical features used from the Z-Alizadeh Sani dataset

| Feature name | Range |
| --- | --- |
| FBS (fasting blood sugar) (mg/dl) | 62–400 |
| Age (years) | 30-86 |
| LDL (low density lipoprotein) (mg/dl) | 18–232 |
| HDL (high density lipoprotein) (mg/dl) | 15–111 |
| Cr (creatine) (mg/dl) | 0.5–2.2 |
| WBC (white blood cell) (cells/ml) | 3700–18,000 |
| BUN (blood urea nitrogen) (mg/dl) | 6–52 |
| K (potassium) (mEq/lit) | 3.0–6.6 |
| HB (hemoglobin) (g/dl) | 8.9–17.6 |
| Na (sodium) (mEq/lit) | 128–156 |
| PLT (platelet) (1000/ml) | 25–742 |
| BP (blood pressure: mmHg) | 90–190 |
| PR (pulse rate) (ppm) | 50–110 |
| TG (triglyceride) (mg/dl) | 37–1050 |
| Neut (neutrophil) (%) | 32–89 |
| Lymph (Lymphocyte) (%) | 7–60 |
| EF (ejection fraction) (%) | 15–60 |
| ESR (erythrocyte sedimentation rate) (mm/h) | 1–90 |
| Gender | Male, female |
| DM (Diabetes Mellitus) | No, yes |
| HTN (hypertension) | No, yes |
| Obesity | No if BMI <= 25, yes otherwise |
| Smoker (Current smoker) | No, yes |
| Ex_Smoker (Previous smoker) | No, yes |
| FH (family history) | No, yes |
| CRF (chronic renal failure) | No, yes |
| CVA (Cerebrovascular Accident) | No, yes |
| Airway Disease | No, yes |
| Thyroid Disease | No, yes |
| DLP (Dyslipidemia) | No, yes |
| Edema | No, yes |
| Weak peripheral pulse | No, yes |
| Lung Rales | No, yes |
| Systolic murmur | No, yes |
| Diastolic murmur | No, yes |
| Typical Chest Pain | No, yes |
| Dyspnea | No, yes |
| Function class | Normal, high |
| Region with RWMA (regional wall motion abnormality) | Normal, high |
| Atypical | No, yes |
| Nonanginal CP | No, yes |
| Q Wave | No, yes |
| ST Elevation | No, yes |
| ST Depression | No, yes |
| T inversion | No, yes |
| LVH (left ventricular hypertrophy) | No, yes |
| Poor R progression (poor R wave progression) | No, yes |
| VHD (valvular heart disease) | Normal, mild, severe, moderate |
| CAD (coronary artery disease) | No, yes |

**Table 3:** Definable profiles based on age, gender, and medical conditions [12]

| Profile# | Age Tag | Gender Tag | Medical Condition Tag |
|---|---|---|---|
| $p_1$ | normal (middle-age) | male | normal |
| $p_2$ | normal (middle-age) | female | normal |
| $p_3$ | high (old) | male | normal |
| $p_4$ | high (old) | female | normal |

**Table 4**: Normal ranges extracted for the profile-based discretization of numerical features

| Feature name | Profile Num | Low | Normal | High |
|---|---|---|---|---|
| FBS2 | $p_1, p_2, p_3, p_4$ | <60 | 60-99 | >99 |
| ESR2 | $p_1, p_3$ | - | ESR <= age/2 | ESR > age/2 |
|  | $p_2, p_4$ | - | ESR <= age/2 + 5 | ESR > age/2 +5 |
| Age2 | $p_1, p_3$ | - | <=45 | >45 |
|  | $p_2, p_4$ | - | <=55 | >55 |
| Cr2 | $p_1, p_3$ | <0.75 | 0.75-1.2 | >1.2 |
|  | $p_2, p_4$ | 0.65 | 0.65-1 | >1 |
| LDL2 | $p_1, p_2, p_3, p_4$ | - | <=130 | >130 |
| HDL2 | $p_1, p_2, p_3, p_4$ | <40 | >=40 | - |
| WBC2 | $p_1, p_2, p_3, p_4$ | <4000 | 4000-10000 | >10000 |
| BUN2 | $p_1, p_2, p_3, p_4$ | <8 | 8-21 | >21 |
| HB2 | $p_1, p_3$ | <13.5 | 13.5-17.5 | >17.5 |
|  | $p_2, p_4$ | <12 | 12-16 | >16 |
| K2 | $p_1, p_2, p_3, p_4$ | <3.4 | 3.4-5.3 | >5.3 |
| Na2 | $p_1, p_2, p_3, p_4$ | <137 | 137-147 | >147 |
| PLT2 | $p_1, p_2, p_3, p_4$ | <150 | 150-399 | >399 |
| BP2 | $p_1, p_2, p_3, p_4$ | <90 | 90-140 | >140 |
| PR2 | $p_1, p_2, p_3, p_4$ | <60 | 60-100 | >100 |
| TG2 | $p_1, p_2, p_3, p_4$ | - | <=200 | >200 |
| Neut2 | $p_1, p_2, p_3, p_4$ | <46 | 46-78 | >78 |
| Lymph2 | $p_1, p_2, p_3, p_4$ | <18 | 18-52 | >52 |
| EF2 | $p_1, p_2, p_3, p_4$ | <=50 | >50 | - |

*4.2. Results of the proposed CAD diagnosis method on the Z-Alizadeh Sani dataset*

This section presents the results of the last phase of the method, i.e. classification with feature selection. The performance of all of the constructed models was evaluated using the 10-fold cross-validation method (in each of the 10 stages, 90% of the data were used for training and the remaining 10% were used for testing). The results of the proposed method on the Z-Alizadeh Sani dataset are reported in Table 5. In this table, the first column gives the Min_sup values used the Apriori algorithm (Phase I, Step 3.1.4), the second column is the number of features extracted from the Z-Alizadeh Sani dataset after executing the Apriori algorithm with the corresponding Min_sup value, the third column is the number of features selected after classification with GA+SVM, and the next four columns show the accuracy, sensitivity, specificity, and AUC of each model. Naturally, a new dataset was created for each Min_sup value (see Section 3.1.4). Each row of this table belongs to the CAD diagnosis model that is built with the corresponding parameters. As these results show, the model built with Min_sup=0.033 gave the best results. From this, it can be inferred that the higher the Min_sup value, the lower the performance of the method (accuracy,sensiticity, specificity and AUC). In other words, the value of Min_sup has a great impact on the performance of the proposed method.

Since the Z-Alizadeh Sani dataset is one of the best datasets in the field of CAD, it has been used in many previous studies. In Table 6, the results of the best model made by the proposed method are compared with the results of previous studies on the same dataset in terms of four criteria: accuracy, sensitivity, specificity, and AUC. As these results demonstrate, the proposed method outperforms all of the methods used in previous credible studies. It can be seen that the use of PBBFE in the proposed method has resulted in much higher diagnosis accuracy than the use of single features as in [21]. Also, the proposed method has about 0.059 better AUC than the powerful method introduced in [35].

Table 5: Performance of the proposed method on the datasets built from the Z-Alizadeh Sani dataset for different Min_sup values

| min support | num.of.extracted.features | num.of.selected.features | Accuracy (%) | Sensitivity (%) | Specificity (%) | AUC |
|---|---|---|---|---|---|---|
| **0.033** | **16382** | **344** | **98.35** | **100** | **94.25** | **0.979** |
| 0.05 | 5815 | 307 | 98.02 | 99.54 | 94.25 | 0.971 |
| 0.075 | 2565 | 298 | 97.69 | 99.07 | 94.25 | 0.969 |
| 0.1 | 1261 | 351 | 97.36 | 99.07 | 93.1 | 0.965 |
| 0.2 | 223 | 180 | 91.75 | 93.52 | 87.36 | 0.908 |
| 0.4 | 23 | 23 | 87.13 | 90.74 | 78.16 | 0.866 |
| 0.5 | 12 | 12 | 85.81 | 89.35 | 77.01 | 0.837 |

Table 6: Performance comparison of the proposed method with previous methods on the Z-Alizadeh Sani dataset

| Method | Accuracy (%) | Sensitivity (%) | Specificity (%) | AUC |
|---|---|---|---|---|
| SMO (with cost matrix) - Alizadehsani [19] | 92.09 | 97.22 | 79.31 | NR* |
| CART – Ghiasi [37] | 92.41 | 98.61 | 77.01 | NR |
| SMO (without feature creation) - Alizadehsani [21] | 92.74 | 95.37 | 86.21 | NR |
| SMO (with feature creation) - Alizadehsani [21] | 94.08 | 96.30 | 88.51 | NR |
| NE-Nu-SVC + Feature Selection +multi-step balancing – Abdar [36] | 94.66 | 94.70 | NR | NR |
| GA for Tuning ANN - Arabasadi [30] | 93.85 | 97 | 92 | NR |
| Feature Engineering - Alizadehsani [35] | 96.4 | 100 | 88.1 | 0.92 |
| FAMD + BBA + RF-ET – Gupta [38] | 97.37 | 98.15 | 95.45 | NR |
| **Our Proposed Method (PBBFE+GA+SVM)** | **98.35** | **100** | **94.25** | **0.979** |

*NR: not reported

Figure 2 shows the ROC curve of the best model of Table 5. As the figure shows, the AUC of the proposed model is approximately 0.98, indicating its excellent performance.

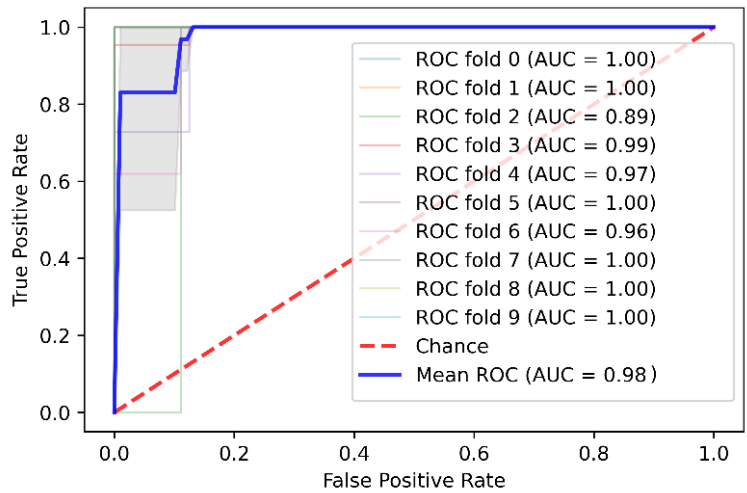

**Fig. 2.** ROC curve of the best model obtained from the proposed method

## 5. Discussion and performance estimation on other datasets

This paper presented a new feature extraction technique to improve the CAD diagnosis accuracy. In the proposed CAD diagnosis method, after the said feature extraction, GA is used to identify the best features among the set of extracted features and an SVM classifier is used to conduct classification based on these features. The method was tested on the Z-Alizadeh Sani dataset.

However, to further evaluate its performance, the proposed method was also applied to two other well-known datasets called Cleveland and Hungarian, which belong to UCI (University of California Irvine) Machine Learning Repository [43]. The Cleveland dataset has 303 records with 14 features and the Hungarian dataset has 294 records with 14 features. Table 7 shows the test results of the proposed method on the Cleveland dataset. As shown in this table, the proposed method produced the best results, achieving an accuracy, sensitivity, and specificity of 95.38%, 94%, and 97%, respectively. As the results of-

**Table 7**: Results of the proposed method on the Cleveland dataset

| Method | Accuracy (%) | Sensitivity | Specificity |
|---|---|---|---|
| FOIL - Das [15] | 64.00 | NR | NR |
| MLP+BP - Das [15] | 65.60 | NR | NR |
| SVM+PCA - Babaoglu [17] | 79.17 | NR | NR |
| SVM | 83.16 | 0.79 | 0.86 |
| Naïve Bayes | 83.50 | 0.82 | 0.85 |
| ANN | 77.55 | 0.72 | 0.82 |
| AIRS [44] | 84.50 | NR | NR |
| ANN - Olaniyi [27] | 85 | NR | NR |
| ANN - Das [15] | 89.01 | 0.81 | 0.96 |
| Weighted fuzzy rules [45] | 86.35 | 0.76 | 0.49 |
| Feature Engineering - Alizadehsani [35] | 93.06 | 0.94 | 0.92 |
| **Our Proposed Method (PBBFE+GA+SVM)** | **95.38** | **0.94** | **0.97** |

**Table 8**: Results of the proposed method on the Hungarian dataset

| Method | Accuracy (%) | Sensitivity | Specificity |
|---|---|---|---|
| Weighted fuzzy rules [45] | 49.93 | 0.40 | 0.74 |
| Naïve Bayes | 80.95 | 0.7 | 0.86 |
| SVM | 82.31 | 0.68 | 0.89 |
| Neural Network | 78.91 | 0.63 | 0.87 |
| RBF | 80.95 | 0.67 | 0.88 |
| RFC - Alizadehsani [35] | 83.20 | 0.85 | 0.77 |
| Feature Engineering - Alizadehsani [35] | 88.77 | 0.88 | 0.89 |
| **Our Proposed Method (PBBFE+GA+SVM)** | **89.12** | **0.9** | **0.89** |

Table 8 show, the proposed method also outperformed all previous methods on the Hungarian dataset, achieving an accuracy, sensitivity, and specificity of 89.12%, 90%, and 89%, respectively. This superiority of the proposed method over previous methods on all datasets can be attributed to:

1) Defining profiles for patients and determining the normal range of each numerical feature based on these profiles (profile-based partitioning);
2) Using the PBBFE technique for feature extraction;
3) High discriminative power and interpretability of extracted binary features (see Appendix A);
4) Not using complex computation procedures for classification (e.g. building kernels with complex mathematical methods).

## 6. Conclusion

This paper presented a two-phase method for CAD diagnosis. In the first phase of the method, which is called PBBFE, the goal is to extract frequent itemsets and use them to improve classification performance. In PBBFE, to increase the accuracy of frequent itemset extraction, each patient is assigned a profile based on age, gender, and medical condition, the normal range of numerical features for each profile is determined, and all numerical features are discretized accordingly. After discretization, all features undergo a binarization process. Then, frequent itemsets are extracted by the Apriori algorithm and added to the original dataset as features. In the second phase, GA is used to identify the best subset of features and then SVM is used for classification. To evaluate the proposed method, it was tested on the Z-Alizadeh Sani dataset, which is one of the most comprehensive and up-to-date datasets related to CAD, was used. The performance evaluation of the proposed method on the Z-Alizadeh Sani dataset and several other datasets showed that the PBBFE technique increased the accuracy of CAD diagnosis. On these datasets, the proposed method performed better than all methods used in the literature.

## Appendix A

Information Gain is a popular feature selection (feature ranking) method, which uses a concept known as entropy. For each feature, the higher the entropy is, the more irregular is the data, which causes the feature to have less information gain (less discriminative power) and vice versa. Thus, features with lower entropy and therefore higher information gain are more valuable. Entropy and information gain of a feature can be calculated by Equations A-1 and A-2, respectively.

$$Entropy(t) = -\sum_{j} p(j|t) \log_2 p(j|t) \tag{A-1}$$

$$InfoGain = Entropy(p) - \sum_{i=1}^{k} \frac{n_i}{n} Entropy(i) \tag{A-2}$$

Table A-1 gives the list of 20 features of the dataset created with Min_sup=0.033 that have the highest information gain. Among the features of this table, one can see one (1-itemset), two (2-itemset), three (3-itemset), and four-dimensional features (4-itemset). As shown in Table A-1, the best feature in terms of information gain is "Typical Chest Pain", and the next best features in this respect are "EF2^Typical Chest Pain", "Age2^Typical Chest Pain", and "Age2^EF2^Typical Chest Pain" in that order.

**Table A-1**: List of top 20 features of the dataset built with min_sup=0.03 in terms of information gain

| Feature Name | Information Gain |
|---|---|
| Typical Chest Pain | 0.231 |
| EF2^Typical Chest Pain | 0.201 |
| Age2^Typical Chest Pain | 0.197 |
| Age2^EF2^Typical Chest Pain | 0.161 |
| Typical Chest Pain^VHD | 0.15 |
| HTN^Typical Chest Pain | 0.145 |
| EF2^Typical Chest Pain^VHD | 0.134 |
| Age2^Typical Chest Pain^VHD | 0.129 |
| EF2^HDL2^Typical Chest Pain | 0.128 |
| EF2^Obesity^Typical Chest Pain | 0.127 |
| FBS2^Typical Chest Pain | 0.122 |
| Age2^HTN^Typical Chest Pain | 0.12 |
| Atypical | 0.12 |
| Obesity^Typical Chest Pain | 0.118 |
| Age2^EF2^Typical Chest Pain^VHD | 0.116 |
| Age2^HDL2^Typical Chest Pain | 0.113 |
| EF2^FBS2^Typical Chest Pain | 0.11 |
| HDL2^Typical Chest Pain | 0.11 |
| EF2^HTN^Typical Chest Pain | 0.11 |
| Region with RWMA2^Typical Chest Pain | 0.106 |